\def\ARXIVVERSION{1}
\documentclass{article}

\newif\ifarxivversion
\ifdefined\ARXIVVERSION
  \arxivversiontrue
\else
  \arxivversionfalse
\fi

\ifdefined\pdfobjcompresslevel
\fi

\usepackage{iclr2027_conference,times}

\usepackage{amsmath,amsfonts,bm}

\def\eqref#1{equation~\ref{#1}}

\def\1{\bm{1}}

\DeclareMathAlphabet{\mathsfit}{\encodingdefault}{\sfdefault}{m}{sl}
\SetMathAlphabet{\mathsfit}{bold}{\encodingdefault}{\sfdefault}{bx}{n}

\usepackage{amssymb}
\usepackage{booktabs}
\usepackage{graphicx}
\usepackage{longtable}
\usepackage{tabularx}
\usepackage{microtype}
\usepackage{url}
\usepackage{xcolor}
\usepackage{enumitem}
\usepackage{float}
\usepackage{placeins}
\makeatletter
\newcommand{\eabNeedspaceFallback}[1]{%
  \par\begingroup
  \setlength{\dimen@}{#1}%
  \vskip\z@\@plus\dimen@
  \penalty-100\vskip\z@\@plus-\dimen@
  \vskip\dimen@\penalty9999\vskip-\dimen@
  \vskip\z@skip\endgroup
}
\makeatother
\IfFileExists{needspace.sty}{
  \usepackage{needspace}
}{
  \let\Needspace\eabNeedspaceFallback
}
\usepackage{array}
\usepackage{hyperref}
\hypersetup{hidelinks}
\AtBeginEnvironment{thebibliography}{\raggedright\interlinepenalty=10000}

\newcommand{\papertitle}{The Evidence Anchoring Boundary: Where Visual Interventions Matter and When They Help in Multimodal LLMs}
\title{\papertitle}

\ifarxivversion
\newcommand{\arxivpdfauthor}{Qiji Zhou, Junyi Hu, Peiyu Wang, Yue Zhang}
\author{%
  Qiji Zhou\textsuperscript{1,*}\quad
  Junyi Hu\textsuperscript{1,2,*}\quad
  Peiyu Wang\textsuperscript{3}\quad
  Yue Zhang\textsuperscript{1,\textdagger}\\[4pt]
  \normalfont\textsuperscript{1}Westlake University
  \quad
  \textsuperscript{2}Duke University
  \\
  \normalfont\textsuperscript{3}Institute of Automation, Chinese Academy of Sciences\\[3pt]
  \normalfont\footnotesize
  \href{mailto:zhouqiji@westlake.edu.cn}{\texttt{zhouqiji@westlake.edu.cn}}\quad
  \href{mailto:junyi.hu@duke.edu}{\texttt{junyi.hu@duke.edu}}\quad
  \href{mailto:yue.zhang@wias.org.cn}{\texttt{yue.zhang@wias.org.cn}}
}

\newcommand{\arxivtitlefootnotes}{%
  \AddToShipoutPictureFG*{%
    \AtPageLowerLeft{%
      \put(\LenToUnit{1.5in},\LenToUnit{0.66in}){%
        \rule{12pc}{0.4pt}%
      }%
      \put(\LenToUnit{1.5in},\LenToUnit{0.48in}){%
        \makebox[0pt][l]{\footnotesize
          \textsuperscript{*}Equal contribution.\quad
          \textsuperscript{\textdagger}Corresponding author.}%
      }%
    }%
  }%
}

  \iclrfinalcopy
  \hypersetup{
    pdftitle={\papertitle},
    pdfauthor={\arxivpdfauthor},
    pdfsubject={arXiv preprint},
    pdfkeywords={multimodal large language models, hallucination mitigation, visual intervention}
  }
\else
  \author{Anonymous Authors}
  \hypersetup{pdftitle={\papertitle},pdfauthor={Anonymous Authors}}
\fi

\begin{document}
\maketitle
\ifarxivversion
  \fancyhead{}
  \renewcommand{\headrulewidth}{0pt}
  \fancyfoot[C]{\thepage}
  \arxivtitlefootnotes
\fi

\begin{abstract}
Visual interventions provide a training-free way to reduce hallucinations in
multimodal large language models. However, an intervention may have little
leverage at some layers and may damage a correct answer even when it changes
the output. We investigate where visual interventions retain leverage and
when they improve predictions. First, we apply the same visual-residual edit
at successive decoder layers, matching its RMS strength across depth. Across six
checkpoints from three model families, we observe a persistent drop in
intervention effect and define its operational onset for this edit as the
\textbf{Evidence Anchoring Boundary (EAB)}.
With boundaries fixed on calibration queries, held-out effects on disjoint
evaluation queries before EAB are $13.78\times$--$26.10\times$ larger than those after it.
Comparison with intermediate readouts shows that the edit can lose leverage
before answer preferences stabilize.
Second, we compare counterfactual subtraction across presence, attribute, and
relation questions. Subtraction can correct a misleading preference, leave a
decision unchanged, or remove a preference supporting the correct answer.
Based on these observations, we develop calibrated selective routing, which
preserves the original prediction unless calibration supports an intervention
for the question type. It improves full-set attribute accuracy by
$1.56$--$1.81$ percentage points across three checkpoints, while universal
subtraction is neutral or harmful.
\end{abstract}

\section{Introduction}

\label{sec:intro}

Multimodal large language models (MLLMs) answer questions about objects,
attributes, and relations in an image. However, their answers can include
details that the image does not support, resulting in multimodal
hallucinations
\citep{rohrbach2018object,li2023pope,wang2023amber,guan2024hallusion,he2026cosimue}.
To understand these errors, previous studies examine where models attend,
what their intermediate states predict, and how internal interventions
affect their outputs
\citep{zhang2025look,belrose2023tuned,shi2025vfl,yang2025modular}.
These measurements capture different aspects of visual processing.
A layer may attend to an object or support an identifiable answer preference,
yet editing its visual representation may have little effect on the final
prediction. How that effect changes with decoder depth remains an
important question for designing interventions.

\begin{figure}[t]
    \centering
    \includegraphics[width=0.95\linewidth]{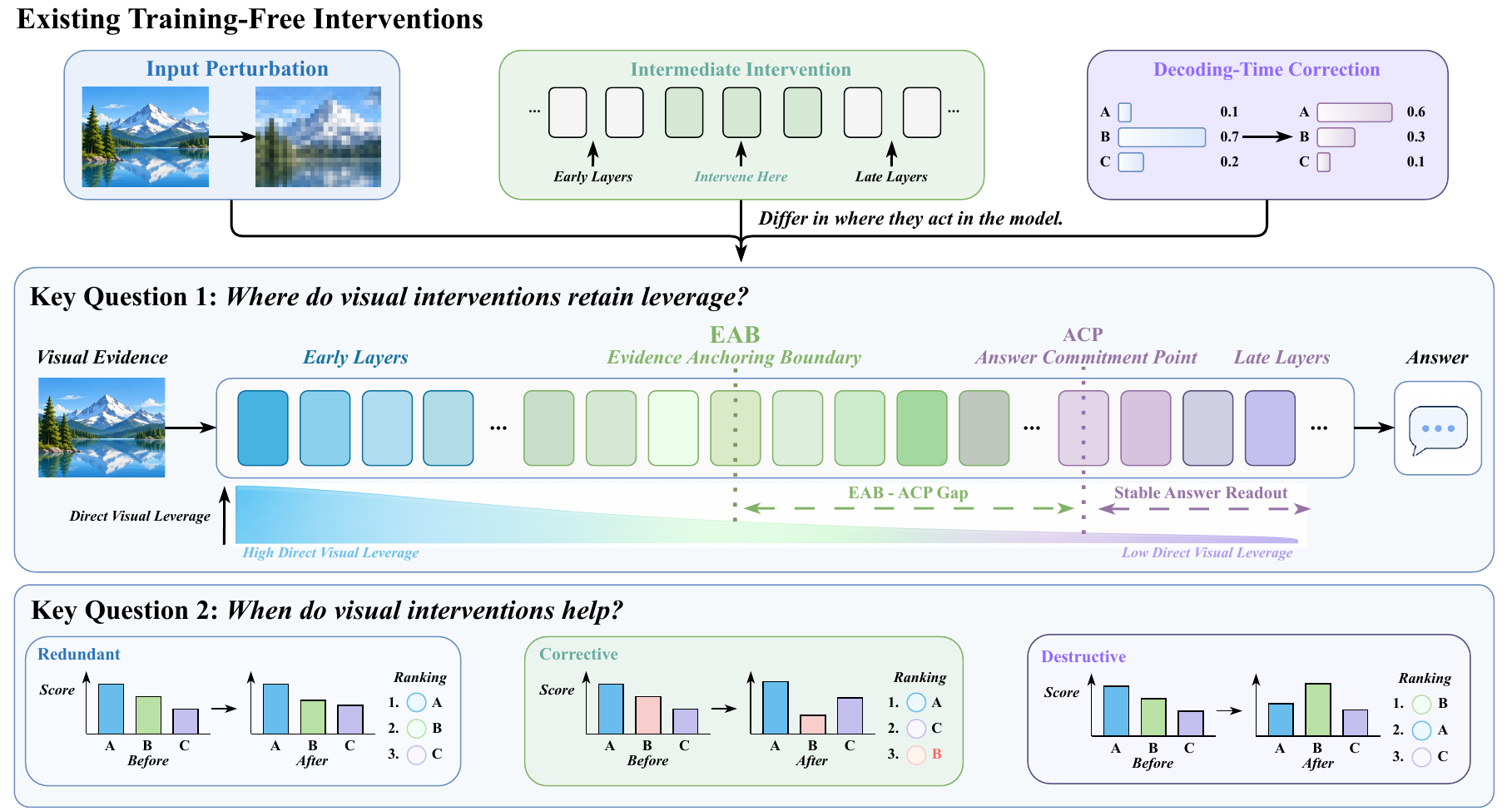}
    \par\smallskip
    \caption{\textbf{From intervention leverage to selective visual intervention.}
Top: training-free intervention sites.
Middle (Q1): the \textbf{Evidence Anchoring Boundary (EAB)} marks where
the tested visual-residual edit becomes persistently weak under equal-RMS perturbations
across depth, potentially before answer stabilization.
Bottom (Q2): an edit can be corrective, redundant, or destructive depending
on the preference it removes, motivating calibrated selective intervention.}
    \label{fig:eab_overview}
\end{figure}

Training-free methods address hallucinations by modifying visual inputs,
intermediate representations, or decoding behavior
(Figure~\ref{fig:eab_overview}, top). They obtain correction signals from
alternative visual conditions, intermediate states, attention patterns, or
decoding statistics during inference. For example, contrastive methods
compare predictions under different visual or instruction conditions, while
other methods measure image dependence or adjust attention
\citep{leng2024vcd,wang2024icd,favero2024m3id,huang2024opera,li2026attentiontracing}.
Layer-selective methods also use internal signals to locate promising
intervention sites \citep{chen2026vihd,shi2025vfl}.
For intervention design, identifying such a site leaves two decisions:
whether a specified edit still has an appreciable effect there, and whether
the resulting change improves the answer for the target question type.

Given the above observations, we investigate two questions:
\textbf{where can a visual intervention still change the final answer (Q1),
and when does that change improve the answer (Q2)?}
First, we apply the same visual-residual edit at successive decoder layers,
matching the root-mean-square (RMS) strength of every edit
(Section~\ref{sec:eab}). This control separates the position of an edit
from its magnitude. Across six checkpoints in three model families,
including 32B and 34B models, we observe a persistent low-effect region.
We call the onset of this low-effect region the \emph{Evidence Anchoring Boundary (EAB)}.
We select one operational boundary per checkpoint on calibration queries, freeze it,
and evaluate the pre-boundary and post-boundary contrast on disjoint queries:
under the frozen rule, pre-EAB effects are \(13.78\times\)--\(26.10\times\) larger than
post-EAB effects.
Here, EAB characterizes the specified edit and boundary rule.

We then compare EAB with the layer where intermediate answer preferences
stabilize, the \emph{Answer Commitment Point (ACP)}. The effect drop can
precede readout stabilization: the model continues to develop its answer
preference after the tested visual edit has lost most of its influence.
A readable intermediate answer and an effective editing site need not
occur at the same depth.

Second, we compare counterfactual subtraction across object-presence,
attribute, and relation questions (Section~\ref{sec:geometry}). We find that
the outcome depends on the relative preference removed from competing
answers. For a yellow bus, removing a preference for \emph{red} can correct
the answer, whereas removing a preference for \emph{yellow} can damage it.
Our measurements establish how these cases occur across tasks: subtraction
improves misleading-attribute subsets, adds little consistent presence-error
information beyond the original confidence, and can hurt direct-contact
relations. The same operation can therefore, in the evaluated settings, be
corrective, redundant, or destructive.

Based on these findings, we develop a \emph{calibrated selective router}
(Figure~\ref{fig:verifier_pipeline}).
Specifically, we first use the question type to determine which
interventions are eligible. We then use calibration data to decide whether
to apply an intervention, preserving the original prediction by default. Experiments
show that universal subtraction changes full-set attribute accuracy by
$-0.42 / -4.92 / -0.33$ points across three checkpoints, whereas routing
improves it by $1.56 / 1.81 / 1.67$ points.
Our main contribution is a controlled characterization of intervention
depth and value: an equal-strength edit can lose influence before answer
readouts stabilize, and subtraction can remove a preference supporting
the correct answer. Selective routing shows how these findings guide
correction.

\section{Related Work}

\label{sec:related}

\textbf{Training-free hallucination mitigation.}
Previous work reduces hallucinations by changing how visual information
contributes to generation during inference
\citep{huang2026chasd,yin2026cec}.
VCD compares predictions from original and distorted images, while ICD
perturbs instructions \citep{leng2024vcd,wang2024icd}. Other methods use
masking, retrieval, or counterfactual inputs to reduce unsupported
preferences \citep{liu2024paying,chen2025ict,li2026dive}.
M3ID measures image dependence, DeCo uses intermediate predictions, and
OPERA applies decoding penalties and retrospection
\citep{favero2024m3id,wang2025deco,huang2024opera,li2026attentiontracing}.
Our counterfactual scores follow this training-free setting. We study the
conditions under which subtraction improves or damages a prediction across
question types and different counterfactual preference structures, and
compare selective use with applying the same subtraction everywhere.

\textbf{Visual localization and layer-selective intervention.}
\citet{zhang2025look} use attention and gradient maps to guide visual
interventions for small details. VIHD identifies decoder windows with high
visual attention, masks selected visual tokens, and uses the resulting
response distributions to detect hallucinations in medical VQA
\citep{chen2026vihd}. These studies establish useful intervention sites
from spatial or layer-specific signals. We instead measure the final-output
effect of a fixed edit at every depth, equalize its RMS strength, and test
a calibration-selected low-effect region on independent queries.
The distinction is between locating a promising site and measuring how
the influence of a specified edit persists across depth for a fixed
intervention operator.

\textbf{Intermediate readouts and causal influence.}
Logit-lens methods project intermediate hidden states into vocabulary space;
tuned lenses learn translators to improve these predictions
\citep{belrose2023tuned}. Causal tracing and activation patching measure how
changing internal representations affects the final output
\citep{meng2022rome,yang2025modular}. Multimodal studies further investigate
visual tokens, attention, and information flow across layers
\citep{shi2025vfl,kang2025visualsink,yang2025modular,chen2026yard,xian2026lad}.
We bring these two measurements together by comparing direct-edit effects
with answer-readout stabilization under the same checkpoint and query
design. EAB summarizes the former trajectory; ACP summarizes the latter.
Their separation shows why intermediate answer availability alone does
not determine when a particular visual edit remains effective, even when
those readouts continue to evolve downstream.

\textbf{Multimodal hallucination evaluation.}
POPE evaluates object presence, AMBER includes attributes and relations,
and HallusionBench tests visually confounding questions and prompt
presuppositions \citep{li2023pope,wang2023amber,guan2024hallusion}.
CHAIR measures object hallucination in generated descriptions
\citep{rohrbach2018object}. We use claim-level comparisons to distinguish
presence, attribute, and relation effects, and report targeted generation
and transfer separately. This separation makes clear which findings concern
binary decisions and which extend to generated answers, while keeping
task-specific evidence distinct from broader claims.

\section{Task and Experimental Setup}
\label{sec:setup}

Both experiments use binary visual claims. Q1 varies the layer of a fixed
edit and measures its final-output effect. Q2 compares original and
counterfactual scores to measure whether subtraction improves the decision.
The shared format keeps intervention effect and task accuracy distinct.

\subsection{Task formulation}
We study the model's preference for a binary visual claim before and after
a controlled visual intervention. For an image of a yellow bus, for example,
``the bus is yellow'' and ``the bus is red'' are competing attribute claims.
An object-presence claim can be evaluated on its own; attribute and relation
claims can also be compared with alternatives about the same image. We call
the claims associated with one image a \emph{claim set},
\(\mathcal{C}=\{c_1,\ldots,c_K\}\). Each \(c_i\) is tested with the same
Yes/No answer format to keep answer scoring directly comparable. For an
image \(x\), we compute
\begin{align}
M(c_i) &= \log p_\theta(\mathrm{Yes}\mid x,c_i)-\log p_\theta(\mathrm{No}\mid x,c_i), \label{eq:margin}\\
M_{\mathrm{cf}}(c_i) &= \log p_\theta(\mathrm{Yes}\mid \bar{x},c_i)-\log p_\theta(\mathrm{No}\mid \bar{x},c_i), \nonumber\\
G(c_i) &= M(c_i)-M_{\mathrm{cf}}(c_i), \label{eq:visual_increment}
\end{align}
Here \(M\) is the original Yes/No log-odds: a positive value favors Yes.
The counterfactual condition \(\bar{x}\) denotes an intervention on
visual-token representations, with the prompt and claim held fixed.
\(M_{\mathrm{cf}}\) is the resulting margin, and \(G\) measures the change
from that condition to the original image. Using \(G\) as the answer score
therefore subtracts the preference that remains under the counterfactual.
That preference can reflect the prompt, model knowledge, and information
retained by the chosen ablation; \(M_{\mathrm{cf}}\) is not assumed to
isolate a pure language prior.
For competing claims, a larger score indicates a stronger preference. Let
\(\Delta Q_{ij}=Q(c_i)-Q(c_j)\), for \(Q\in\{M,M_{\mathrm{cf}},G\}\). Then
\begin{equation}
\Delta G_{ij}=\Delta M_{ij}-\Delta M_{\mathrm{cf},ij}.
\label{eq:shear_identity}
\end{equation}
Thus, the relative counterfactual preference
\(\Delta M_{\mathrm{cf},ij}\) determines how subtraction changes the
comparison between two claims. A positive \(\Delta M_{ij}\) ranks \(c_i\)
above \(c_j\). For untied scores, subtraction reverses their ranking when
\(\Delta M_{ij}\) and \(\Delta G_{ij}\) have opposite signs. The identity specifies the score change,
while the experiments determine whether that change is beneficial.

\subsection{Models, datasets, and evaluation scope}
Our base checkpoints are LLaVA-1.5-7B/13B, Qwen2.5-VL-7B, and
InternVL3.5-8B \citep{liu2023llava,bai2025qwenvl,chen2025internvl35}.
We additionally use LLaVA-NeXT-34B and Qwen2.5-VL-32B to test Q1 at larger
scale. Table~\ref{tab:main_coverage} summarizes the tasks and datasets.
Q2 coverage varies by checkpoint; Appendix~\ref{app:coverage} gives the
evaluated combinations and sample counts.

\begin{table}[H]
\centering
\small
\setlength{\tabcolsep}{5pt}
\caption{\textbf{Main evaluation map.} Q1 uses 120 shared queries per
checkpoint (48 calibration, 72 disjoint evaluation); Q2 coverage varies by
checkpoint as recorded in Appendix~\ref{app:coverage}.}
\label{tab:main_coverage}
\begin{tabular}{@{}lll@{}}
\toprule
Question & Dataset / task & Use in this paper \\
\midrule
Q1: layer-wise effect & Shared sampled queries & EAB and answer-readout comparison \\
Q2: presence & POPE; TVQA-Existence & Confidence-conditioned diagnostics \\
Q2: attributes & AMBER claim pairs & Outcome analysis and routing \\
Q2: relations & AMBER claim pairs & Outcome analysis \\
\bottomrule
\end{tabular}
\end{table}

\subsection{Intervention and metrics}
For Q1, we edit the visual residual contribution at successive decoder
layers and match the root-mean-square strength of the edits across depth.
We propagate each edited state through the remaining decoder and measure
the change in its final output. The boundary is selected on the calibration
split and frozen before measuring the held-out pre-/post-boundary contrast.
For Q2, we construct \(\bar{x}\) by mean-ablating visual-token
representations. This counterfactual supplies the preference removed by
subtraction; it is distinct from the equal-RMS depth sweep used for Q1
for comparison.

For presence questions, we compare correct and incorrect Yes predictions
at matched original confidence \(M\). Cross-fitted residual
\(\operatorname{AUC}(G\mid M)\) and incremental log loss measure whether \(G\) provides
additional information about these errors; residual AUC near \(0.5\)
indicates little additional discrimination. For attributes and relations,
we compare pair accuracy under \(M\) and \(G\).
The hard-attribute subset uses answer labels to explain the role of a
misleading counterfactual preference. The router is evaluated on the full
attribute set, with no access to this analysis-only label at test time.

\subsection{Calibration, comparisons, and uncertainty}
Calibration and evaluation use disjoint images, keeping every claim from
one image together. We select the attribute routing rule by five-fold
image-disjoint cross-validation; each image is scored by a rule selected
without it, preventing leakage. Original scoring \(M\), universal subtraction \(G\), and
routing are evaluated on the same images. This comparison tests whether
selection improves a given intervention, with the matched-token ICD/DeCo
audits reported separately in Appendix~\ref{app:secondary_baselines}.

Confidence intervals use paired image-cluster resampling, preserving
dependencies among claims about the same image rather than treating claims
as independent. For Q1, the held-out intervals quantify the effect ratio
conditional on the frozen boundary without reselection.
Appendices~\ref{app:experimental_protocol} and~\ref{app:reproducibility_checklist}
give the split and reproduction protocols;
Appendix~\ref{app:significance_audit} reports the paired tests.

\section{Q1: Where Can a Visual Intervention Matter?}
\label{sec:eab}

We investigate how the effect of a direct visual-residual edit changes
across decoder layers under matched intervention strength, then compare
this change with answer stabilization.

\subsection{A persistent drop in intervention effect}
\begin{figure}[H]
\centering
\includegraphics[width=1.0\linewidth]{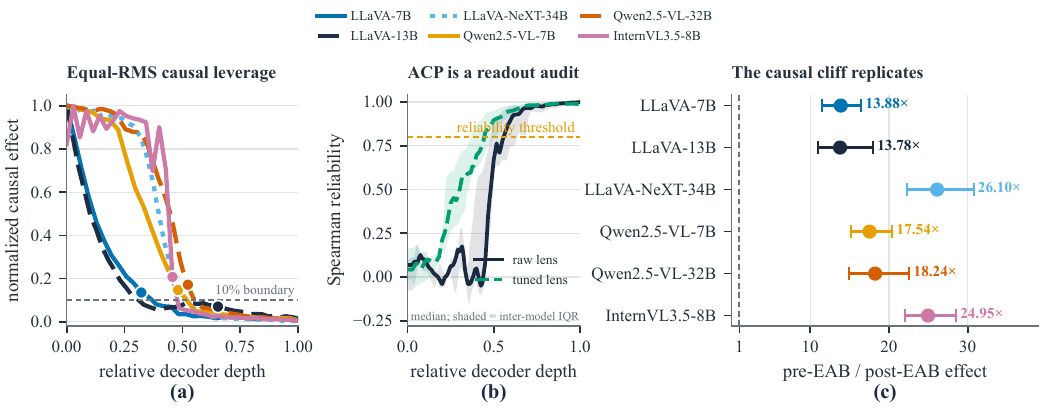}
\caption{\textbf{EAB identifies where the tested direct visual edit loses leverage.}
(a) Checkpoint-level boundaries frozen on calibration queries identify a
persistent low-effect regime on disjoint evaluation queries under the same
equal-RMS strength control. (b)
Answer-readout reliability can continue to increase after this measured drop;
ACP marks answer-readout stabilization. (c) Fully held-out pre-EAB
effects are \(13.78\times\)--\(26.10\times\) post-EAB effects across six
checkpoints.}
\label{fig:eab_discovery}
\end{figure}

Figure~\ref{fig:eab_discovery}a shows the effect of moving the same edit
through six checkpoints. We observe larger effects at earlier layers,
followed by a persistent decrease. We call the onset of this low-effect
region the \emph{Evidence Anchoring Boundary} (EAB). Here, anchoring names
the observed loss of influence of this direct visual edit. The boundary
is an operational summary of its effect trajectory at a given checkpoint,
rather than a universal architectural transition shared across interventions.

Specifically, we edit the state at layer \(\ell\), pass it through the
remaining layers, and measure changes in the final Yes/No margin, answer,
and output distribution. All edits have equal root-mean-square strength.
For original and intervened output distributions \(p\) and
\(p^{(\ell,\mathrm{int})}\), the distributional effect is
\begin{equation}
D_\ell=D_{\mathrm{KL}}\!\left(p\,\|\,p^{(\ell,\mathrm{int})}\right),
\label{eq:layer_effect}
\end{equation}
We locate EAB using calibration queries. We select the earliest layer after
the effect peak at which both the three-layer local mean and the 75th
percentile of the remaining layers fall below \(10\%\) of that peak.
We then fix the boundary and evaluate it on disjoint queries.
Figure~\ref{fig:eab_discovery}c shows that effects before EAB are
\(13.78\times\)--\(26.10\times\) those after it, with every
image-cluster confidence interval excluding one. The larger checkpoints
show the same pattern: \(26.10\times\) [\(22.24,30.74\)] for
LLaVA-NeXT-34B and \(18.24\times\) [\(14.89,22.59\)] for Qwen2.5-VL-32B.
These ratios establish a held-out contrast between early and late effects
under the frozen rule. The reported layer marks the onset of its operational
low-effect region. Appendix~\ref{app:boundary_sensitivity} discusses location
uncertainty and the distinction from a change point.

\subsection{Separating intervention influence from answer stabilization}
To examine whether the drop occurs because the answer has already
stabilized, we compare EAB with intermediate answer predictions.
We obtain raw readouts by applying the final normalization and
language-model head to the answer-position hidden state \(h_\ell\):

\begin{equation}
z_\ell=W_{\mathrm{LM}}\mathrm{LN}_{\mathrm{final}}(h_\ell),
\qquad
m_\ell=z_\ell(\mathrm{Yes})-z_\ell(\mathrm{No}).
\label{eq:intermediate_readout}
\end{equation}
We use raw and group-held-out binary tuned lenses to identify the
\emph{Answer Commitment Point} (ACP), the earliest layer after which
decoded margin signs and candidate rankings remain aligned with the final
prediction \citep{belrose2023tuned,shi2025vfl}.
Figure~\ref{fig:eab_discovery}b shows that the readouts can continue to
improve after EAB. This result separates two processes: intermediate
answer preferences are still developing, while the tested direct visual
edit already has little influence on the final output.
Post-ACP query edits can also change answers
(Appendix~\ref{app:acp_boundary}), supporting the use of ACP as a readout
comparison. Appendix~\ref{app:operator_robustness} describes the operator scope.

The first experiment identifies where the tested edit remains effective.
We next investigate what makes an intervention useful for a given question
beyond causal influence alone.

\section{Q2: When Should a Visual Intervention Be Used?}
\label{sec:geometry}

\begin{figure}[H]
\centering
\includegraphics[width=\linewidth]{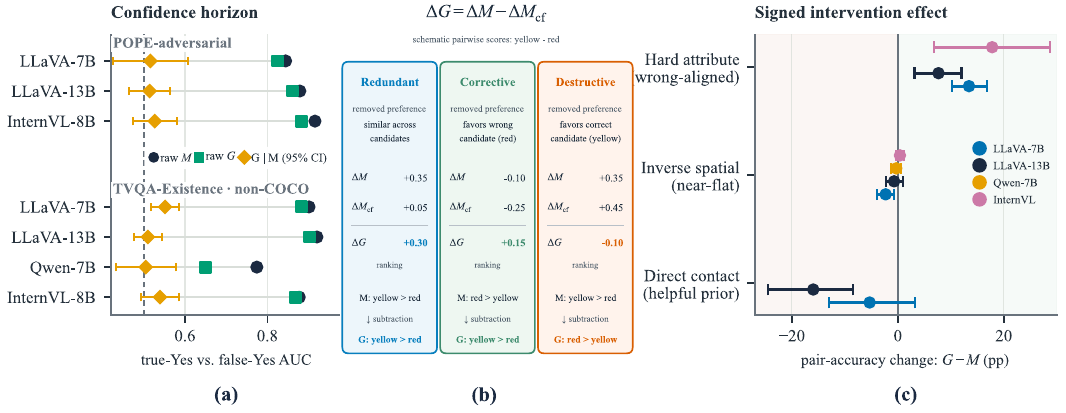}
\caption{\textbf{The removed preference explains three intervention outcomes.}
(a) Presence: \(G\) adds little consistent discrimination between correct
and incorrect Yes predictions after conditioning on \(M\).
(b) Schematic yellow-minus-red score differences illustrate
\(\Delta G=\Delta M-\Delta M_{\mathrm{cf}}\): subtraction preserves,
corrects, or damages the ranking according to the removed preference.
The values are illustrative, not measured.
(c) Observed pair-accuracy changes for hard attributes, inverse-spatial
relations, and direct-contact relations; positive values favor \(G\).
Error bars show 95\% image-cluster confidence intervals, with uncertainty
estimated over image-level dependence structures.}
\label{fig:claim_geometry}
\end{figure}

Why does counterfactual subtraction improve some answers but damage others?
For the yellow-bus example (Section~\ref{sec:setup}), removing a preference
for \emph{red} can correct an error; removing a preference for \emph{yellow}
can damage a correct answer. Equal counterfactual scores leave the difference
unchanged for the competing claims considered
(Equation~\ref{eq:shear_identity}).
Figure~\ref{fig:claim_geometry}b illustrates the three ranking outcomes with
schematic scores. In the corrective case,
\(\Delta G=-0.10-(-0.25)=+0.15\), reversing the preference from red to
yellow. Panels (a) and (c) show the aggregate evidence.

\subsection{Limited benefit from subtraction}
For object presence, we compare correct and incorrect Yes predictions
to test whether \(G\) adds information beyond the original confidence \(M\).
In Figure~\ref{fig:claim_geometry}a, residual \(\operatorname{AUC}(G\mid M)\) on POPE-adversarial is \(0.516\), \(0.514\), and
\(0.526\) for LLaVA-7B, LLaVA-13B, and InternVL. The intervals cover chance,
and incremental log-loss gains are unresolved.
The non-COCO TVQA-Existence values are \(0.552/0.510/0.505/0.539\)
across the four base checkpoints. These results show little consistent
additional error discrimination after accounting for \(M\) across benchmarks.

We also examine inverse-spatial claim pairs. Their counterfactual scores
are similar across candidates, so subtraction changes the relative scores
little. Accuracy changes range from \(-2.3\) to \(+0.4\) points
(Figure~\ref{fig:claim_geometry}c, middle); the LLaVA-7B decline is
statistically resolved (Appendix~\ref{app:significance_audit}).
Appendices~\ref{app:attention_diagnostic} and~\ref{app:qualitative_presence}
give the related attention diagnostics and examples for completeness.

\subsection{When the removed preference is misleading or task-relevant}
Using answer labels for analysis only, we identify AMBER attribute pairs
whose counterfactual favors the incorrect candidate. On this hard
subset, subtraction improves pair accuracy across checkpoints by \(13.44\),
\(7.68\), and \(17.81\) points for LLaVA-7B, LLaVA-13B, and InternVL.

However, the full attribute sets contain both corrective and destructive
cases under the same intervention. Applying \(G\) to every pair changes
accuracy by \(-0.42\), \(-4.92\), and \(-0.33\) points. For direct-contact
relations, the changes are \(-5.32\) and \(-15.96\) points on the two LLaVA
checkpoints. Figure~\ref{fig:claim_geometry}c contrasts these effects:
subtraction can also remove a task-relevant preference when the original
ranking is correct. Table~\ref{tab:category_results} summarizes the results;
Appendix~\ref{app:qualitative_attributes} gives examples.

\begin{table}[!t]
\centering
\scriptsize
\setlength{\tabcolsep}{0.9pt}
\renewcommand{\arraystretch}{1.08}
\caption{\textbf{Subtraction has different effects across question types
for each evaluated checkpoint.}
Presence columns report residual \(\operatorname{AUC}(G\mid M)\); all other columns report
the accuracy change from replacing \(M\) with \(G\), in percentage points.
(Bracketed values are 95\% image-cluster confidence intervals; --- indicates
that the checkpoint--task pair was not evaluated.)}
\label{tab:category_results}
\begin{tabularx}{\textwidth}{@{}>{\raggedright\arraybackslash}p{0.16\textwidth}*{6}{>{\centering\arraybackslash}X}@{}}
\toprule
Checkpoint & \multicolumn{2}{c}{Presence: residual \(\operatorname{AUC}(G\mid M)\)} & \multicolumn{4}{c}{Paired tasks: accuracy change \(G-M\) (percentage points)} \\
\cmidrule(lr){2-3}\cmidrule(l){4-7}
& \shortstack{POPE-\\adversarial} & \shortstack{TVQA\\non-COCO} & \shortstack{Hard attributes\\(misleading)} & \shortstack{Full attributes\\(mixed)} & \shortstack{Direct contact\\(task-relevant)} & \shortstack{Inverse spatial\\(near-flat)} \\
\midrule
LLaVA-1.5-7B & \(.516\) & \(.552\) & \shortstack{\(+13.44\)\\\([10.17,16.89]\)} & \shortstack{\(-0.42\)\\\([-1.64,0.84]\)} & \shortstack{\(-5.32\)\\\([-12.90,3.16]\)} & \(-2.3\) \\
LLaVA-1.5-13B & \(.514\) & \(.510\) & \shortstack{\(+7.68\)\\\([3.15,12.01]\)} & \shortstack{\(-4.92\)\\\([-6.39,-3.36]\)} & \shortstack{\(-15.96\)\\\([-24.47,-8.42]\)} & \(-0.7\) \\
\midrule
Qwen2.5-VL-7B & --- & \(.505\) & --- & --- & --- & \(-0.3\) \\
\midrule
InternVL3.5-8B & \(.526\) & \(.539\) & \shortstack{\(+17.81\)\\\([6.85,28.77]\)} & \shortstack{\(-0.33\)\\\([-2.83,2.17]\)} & --- & \(+0.4\) \\
\bottomrule
\end{tabularx}
\end{table}

\subsection{Calibrated selective routing}
\label{sec:verifier}

The preceding experiments suggest preserving the original prediction
unless the question type and calibration results support a correction.
We implement this rule with \emph{calibrated selective routing}.
First, we assign the question category using dataset metadata or fixed
rules over question form. Second, we restrict the eligible interventions
to those evaluated for that category. We then use calibration data to
select an intervention or retain \(M\) before any evaluation image is ever
scored. Evaluation labels and correctness are excluded from these decisions
to avoid test-time leakage.

\begin{figure}[H]
\centering
\includegraphics[width=\linewidth]{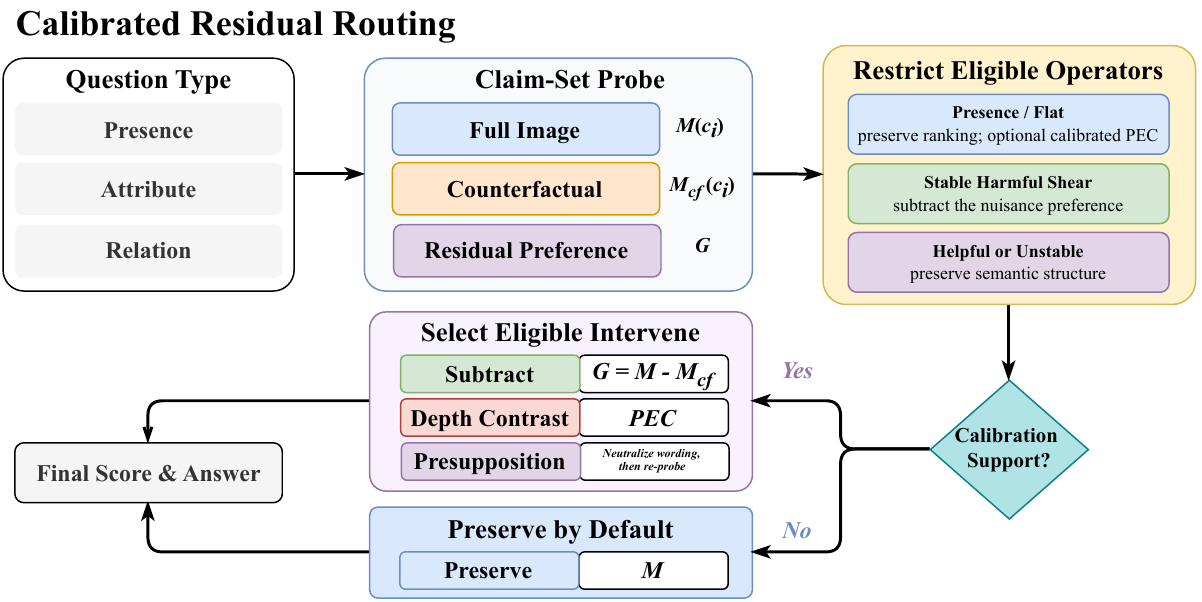}
\caption{\textbf{Calibrated selective routing preserves the original score by default, using only calibration-supported operators for each question category.}
Question category and full-image/counterfactual probes restrict eligible
operators. Calibration supports subtraction for suitable attribute questions
or PEC in its validated presence setting; otherwise the router retains \(M\).
For presupposed questions, neutralization is followed by re-probing before
a score adjustment is selected. All routing rules are fixed before
evaluation, and evaluation labels are excluded from the decision.}
\label{fig:verifier_pipeline}
\end{figure}

Figure~\ref{fig:verifier_pipeline} summarizes this sequence: category and
probes restrict the candidate operators, and calibration determines whether
to adjust the score or preserve \(M\) by default.

For attributes, the router measures how strongly the counterfactual favors
one claim over another and uses the score to interpolate smoothly between
preservation and subtraction:
\begin{align}
A(\mathcal C)&=\max_i M_{\mathrm{cf}}(c_i)-\min_i M_{\mathrm{cf}}(c_i),\\
S(c_i)&=(1-\beta)M(c_i)+\beta G(c_i),\qquad \beta\in[0,1].
\label{eq:routed_score}
\end{align}
Here \(A(\mathcal C)\) measures the spread of counterfactual scores.
Setting \(\beta=0\) retains \(M\), while \(\beta=1\) uses \(G\);
intermediate values apply partial subtraction. We select the rule on
calibration images using this spread, the fraction of rankings that
differ under \(M\) and \(G\) (DisFrac), and the consistency of the
counterfactual-difference sign across equivalent prompts and ablations.
Flat, rarely disagreeing, or unstable comparisons retain \(M\).
Stable differences can support subtraction when calibration shows a benefit
and the sign remains consistent across conditions.

For presence questions, \(G\) provides little additional error information,
so the default score remains \(M\). In the calibrated POPE-adversarial
setting, we additionally allow Prior-Echo Cancellation (PEC):
\begin{equation}
\mathbf z_{\mathrm{PEC}}=\mathbf z_L+\alpha\left(\mathbf z_L-\mathbf z_{\ell^\star}\right).
\label{eq:pec}
\end{equation}
This adjustment uses the difference between final logits \(\mathbf z_L\)
and intermediate logits \(\mathbf z_{\ell^\star}\). We select the layer
\(\ell^\star\) and coefficient \(\alpha\) on calibration data, independently
of EAB. EAB identifies effective layers for direct visual-state edits;
PEC adjusts output scores. We use PEC only in its validated presence
setting. Relation questions retain \(M\) unless relation-specific
calibration supports subtraction.

For presupposed questions, we neutralize the wording and re-probe before
applying the calibrated rule, so the decision uses the neutralized query
for scoring (Appendix~\ref{app:prompt_handling}).

We use five-fold image-disjoint cross-validation for attribute routing.
In each fold, we fix the rule before scoring evaluation images, without any
evaluation-time adaptation within folds.
Appendices~\ref{app:image_disjoint_calibration}--\ref{app:prompt_handling}
describe calibration, category assignment, and prompt handling.

\section{Evaluation: Benefits, Ablations, and Scope}
\label{sec:experiments}

\begin{figure}[H]
\centering
\includegraphics[width=0.96\linewidth]{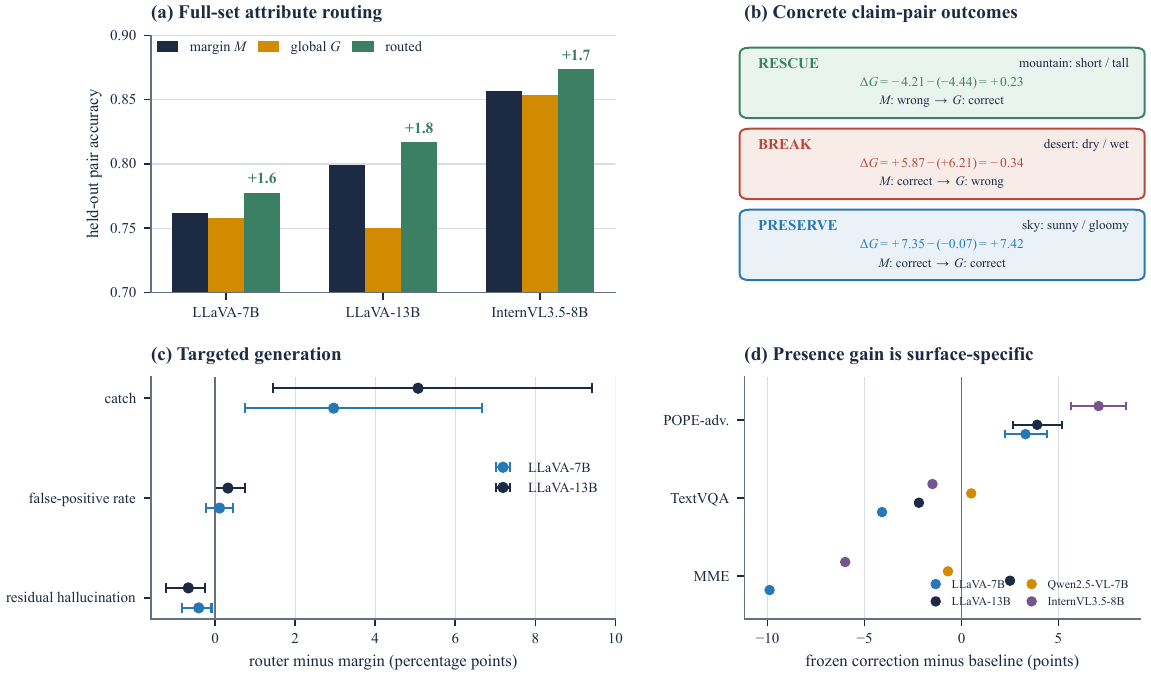}
\caption{\textbf{Selective routing improves attributes; PEC has limited transfer.}
(a) Held-out AMBER attribute accuracy for the original margin \(M\), global
subtraction \(G\), and the routed score. (b) Cached pairs illustrate rescue,
break, and preservation through \(\Delta G=\Delta M-\Delta M_{\mathrm{cf}}\).
(c) Targeted-generation changes on LLaVA-7B/13B, with 95\% image-cluster
bootstrap intervals. (d) Frozen PEC effects across POPE-adversarial, TextVQA,
and MME: PEC improves its calibrated POPE setting but transfers negatively or
inconsistently. Complete results and paired tests are in
Appendices~\ref{app:routing_outcomes}--\ref{app:significance_audit}.}
\label{fig:selective_routing_results}
\end{figure}

We compare the router with the original score \(M\) and universal
subtraction \(G\) to test the benefit of selecting when to intervene
under fixed routing rules.
We then evaluate targeted attribute generation and the transfer of PEC to
other tasks. Figure~\ref{fig:selective_routing_results} summarizes the results.

\Needspace{13\baselineskip}
\begin{samepage}
\subsection{Attribute routing and targeted generation}
Universal subtraction changes full-set attribute accuracy by \(-0.42\),
\(-4.92\), and \(-0.33\) points on LLaVA-7B, LLaVA-13B, and InternVL.
On the same checkpoints, routing improves accuracy by \(1.56\), \(1.81\),
and \(1.67\) points (Figure~\ref{fig:selective_routing_results}a). All three routing
gains have image-cluster confidence intervals above zero
(Appendix~\ref{app:significance_audit}).
These results support selecting when to subtract: the full sets contain
cases where removing the counterfactual preference damages the answer.

We further apply the frozen routing rules to targeted attribute generation.
Hallucination-catch estimates increase by \(2.96\) and \(5.07\) points on
the two LLaVA checkpoints. False-positive changes are small and unresolved,
and residual hallucination decreases
(Figure~\ref{fig:selective_routing_results}c).

\end{samepage}

\Needspace{9\baselineskip}
\subsection{Presence-specific intervention and transfer failures}
Frozen PEC improves POPE-adversarial accuracy by \(3.30\), \(3.90\), and
\(7.07\) points on its three evaluated checkpoints. Its effects on TextVQA
are \(-4.1/-2.2/+0.5/-1.5\) points and on MME are
\(-9.9/+2.5/-0.7/-6.0\) points across LLaVA-7B/13B, Qwen2.5-VL-7B,
and InternVL (Figure~\ref{fig:selective_routing_results}d).
POPE-random changes are only \(+0.8/+0.8/-0.3\) points. We therefore
restrict PEC to its calibrated POPE-adversarial setting rather than
assuming these gains transfer beyond calibration.

\subsection{Selection, calibration, and failure modes}

\begin{table}[H]
\centering
\scriptsize
\setlength{\tabcolsep}{3.0pt}
\renewcommand{\arraystretch}{1.06}
\caption{\textbf{Evidence for selection and task-specific calibration across
the evaluated checkpoints.}
Controlled comparisons and observed failure modes. Effects are
percentage-point changes unless noted. (Attribute and POPE tests are in
Appendix~\ref{app:significance_audit}; direct-contact intervals are in
Table~\ref{tab:category_results}.)}
\label{tab:routing_ablations}
\begin{tabularx}{\textwidth}{@{}>{\raggedright\arraybackslash}p{0.25\textwidth}>{\raggedright\arraybackslash}p{0.29\textwidth}>{\raggedright\arraybackslash}X@{}}
\toprule
Comparison / design choice & In-domain evidence & Failure / implication \\
\midrule
Use \(G\) for every attribute (remove selective routing) & \textit{Routed score minus \(M\)}\par\smallskip
\begin{tabular}[t]{@{}l@{\quad}r@{}}
LLaVA-1.5-7B & \(+1.56\) \\
LLaVA-1.5-13B & \(+1.81\) \\
InternVL3.5-8B & \(+1.67\)
\end{tabular} & \textit{Global \(G\) minus \(M\)}\par\smallskip
\begin{tabular}[t]{@{}l@{\quad}r@{}}
LLaVA-1.5-7B & \(-0.42\) \\
LLaVA-1.5-13B & \(-4.92\) \\
InternVL3.5-8B & \(-0.33\)
\end{tabular} \\
\midrule
Use subtraction across question types & \textit{Presence residual AUC}\par\smallskip
\begin{tabular}[t]{@{}l@{\quad}r@{}}
LLaVA-1.5-7B & \(.516\) \\
LLaVA-1.5-13B & \(.514\) \\
InternVL3.5-8B & \(.526\)
\end{tabular} & \textit{Direct-contact accuracy change}\par\smallskip
\begin{tabular}[t]{@{}l@{\quad}r@{}}
LLaVA-1.5-7B & \(-5.32\) \\
LLaVA-1.5-13B & \(-15.96\)
\end{tabular} \\
\midrule
Apply frozen PEC across tasks & \textit{POPE-adversarial gain}\par\smallskip
\begin{tabular}[t]{@{}l@{\quad}r@{}}
LLaVA-1.5-7B & \(+3.30\) \\
LLaVA-1.5-13B & \(+3.90\) \\
InternVL3.5-8B & \(+7.07\)
\end{tabular} & \textit{Frozen transfer change}\par\smallskip
\begin{tabular}[t]{@{}l@{\;}r@{\;}r@{\;}r@{}}
Checkpoint & \shortstack{POPE-\\random} & TextVQA & MME \\
LLaVA-1.5-7B & \(+0.8\) & \(-4.1\) & \(-9.9\) \\
LLaVA-1.5-13B & \(+0.8\) & \(-2.2\) & \(+2.5\) \\
Qwen2.5-VL-7B & --- & \(+0.5\) & \(-0.7\) \\
InternVL3.5-8B & \(-0.3\) & \(-1.5\) & \(-6.0\)
\end{tabular} \\
\midrule
Retain the stability safeguard & Separate contribution not quantified. & Reject unstable corrections; no isolated accuracy gain is claimed. \\
\bottomrule
\end{tabularx}
\end{table}

Table~\ref{tab:routing_ablations} examines selection and calibration.
Applying subtraction to every attribute loses the routing gains, while
applying PEC beyond its calibrated task can harm performance.
These comparisons support the need to decide where each correction is useful
within its calibrated scope.
Appendices~\ref{app:verifier_details} and~\ref{app:secondary_baselines}
describe calibration and the scope of the secondary baseline comparisons.

\section{Limitations}
\label{sec:discussion}

EAB describes the tested visual-residual edit at each checkpoint, while
ACP measures answer-readout stabilization. The 48/72 calibration/evaluation
split supports the reported effect contrast; a larger, more diverse sample
is needed to estimate how precisely the boundary location generalizes
across settings. The numerical results cover one boundary rule and one primary Q1 operator.
New models, prompts, operators, or task categories require recalibration.
The routing gains are established mainly for binary attribute claims;
broader generation and faithful end-to-end comparisons with external
decoding methods remain open. Additional counterfactual passes create a
compute cost (Appendix~\ref{app:inference_cost}); wall-clock latency and
FLOP overhead remain unmeasured.

\Needspace{10\baselineskip}
\section{Conclusion}
We investigate where visual interventions remain effective and when they
improve answers. The tested direct visual edit shows a persistent effect
drop that can precede answer-readout stabilization. Across question types,
subtraction helps or harms according to the preference it removes.
Calibrated selective routing improves attribute accuracy where universal
subtraction is neutral or harmful. The findings motivate two checks before
applying a visual correction: measure the chosen operator's influence at
the intended depth, and evaluate the resulting decisions on the target
task. Selective routing applies this second check to counterfactual subtraction
in calibrated settings.

\clearpage

\subsection*{AI use statement}

In this work, we used generative AI tools to assist with drafting, revising,
and polishing portions of the manuscript and with writing and debugging a
subset of the code used to generate figures. These tools were not treated as
sources of evidence. The authors reviewed and edited all AI-assisted text,
inspected the AI-assisted plotting code, and verified the resulting figures
against the underlying experimental outputs. The authors take full
responsibility for the final content of this work, including its text, code,
figures, claims, and other artifacts produced with AI assistance.

\subsection*{Ethics statement}

This work evaluates open-source MLLMs on established public
benchmarks and does not involve human-subject experiments or the collection of
new personal or sensitive data. We do not release a new dataset or deploy a
user-facing system. The studied interventions are intended to reduce
unsupported multimodal outputs; however, our transfer results show that they
can degrade performance outside their calibrated settings. We therefore
restrict our claims to the evaluated models and tasks and emphasize calibrated,
default-preserving use. Beyond the general risks associated with MLLM bias,
hallucination, and misuse, we are not aware of additional ethical concerns
specific to this study.

\subsection*{Reproducibility statement}

Section~\ref{sec:setup} specifies the output scores, interventions, evaluated
checkpoints, and claim-specific evaluation scope.
Appendix~\ref{app:experimental_details} maps each claim to its evaluated
model--task cells and records the experimental protocol and reproduction
environment. Appendix~\ref{app:q1_additional} provides the checkpoint-level
evidence for Q1, and Appendix~\ref{app:boundary_robustness} explains the
intervention control and boundary interpretation;
Appendix~\ref{app:q2_additional} provides the diagnostic
checks for Q2; Appendix~\ref{app:verifier_details} specifies the held-out
calibration protocol; and Appendix~\ref{app:extended_evaluation} reports the
extended outcomes, transfer tests, and paired significance tests.
Appendix~\ref{app:qualitative} contains mechanism illustrations that are not
used for tuning or prevalence claims. Together, these records support the
scope, interpretation, and reproduction of the main-text results.



\bibliographystyle{iclr2027_conference}
\bibliography{references}

\clearpage
\appendix

\section{Experimental Coverage and Reproducibility}
\label{app:experimental_details}

We give the model and dataset coverage for Section~\ref{sec:setup},
followed by the data splits, intervention settings, and reproduction
records needed for independent reproduction.

\subsection{Checkpoint and dataset coverage}
\label{app:coverage}
Table~\ref{tab:models_data} lists the experiments used in
Sections~\ref{sec:eab}--\ref{sec:experiments}. Q1 uses the same 120-query
design for all six checkpoints. Q2 uses the four base checkpoints, with
attribute, relation, and downstream experiments evaluated on the listed
subsets. A dash indicates an unevaluated combination.

\begin{table}[H]
\centering
\scriptsize
\setlength{\tabcolsep}{1.8pt}
\renewcommand{\arraystretch}{1.16}
\caption{\textbf{Claim-specific model and dataset coverage.} Every Q1 checkpoint uses 120
queries (48 calibration/72 evaluation); Q2 and deployment claims are restricted
to the populated cells to match their evaluated scope. (POPE counts are
adversarial/random; --- indicates that the checkpoint--task pair was not
evaluated. Statistical audits report their own sample counts in
Table~\ref{tab:paired_significance}. LLaVA-NeXT-34B and Qwen2.5-VL-32B are
Q1-only scale replications.)}
\label{tab:models_data}
\begin{tabular}{@{}>{\raggedright\arraybackslash}p{0.15\textwidth}>{\centering\arraybackslash}p{0.09\textwidth}>{\centering\arraybackslash}p{0.18\textwidth}>{\centering\arraybackslash}p{0.10\textwidth}>{\centering\arraybackslash}p{0.11\textwidth}>{\raggedright\arraybackslash}p{0.29\textwidth}@{}}
\toprule
Checkpoint & \shortstack{Q1 queries\\(EAB/ACP)} & \shortstack{Presence\\claim sets} & \shortstack{Attribute\\pairs} & \shortstack{Relation\\pairs} & Downstream evaluation \\
\midrule
\multicolumn{6}{@{}l}{\textbf{Base checkpoints used beyond Q1}} \\
LLaVA-1.5-7B & 120 & \shortstack{POPE: 3,000/3,000\\TVQA: 2,000} & 2,378 & \shortstack{contact: 94\\spatial: 1,000} & attribute router; targeted generation: 1,040;\newline PEC: 3,000; CHAIR: 500 \\
LLaVA-1.5-13B & 120 & \shortstack{POPE: 3,000/3,000\\TVQA: 2,000} & 2,378 & \shortstack{contact: 94\\spatial: 1,000} & attribute router; targeted generation: 1,077;\newline PEC: 3,000; CHAIR: 500 \\
Qwen2.5-VL-7B & 120 & \shortstack{POPE: 3,000/3,000\\TVQA: 2,000} & --- & spatial: 1,000 & PEC transfer: TextVQA/MME;\newline POPE PEC omitted\newline (two baseline false-Yes cases) \\
InternVL3.5-8B & 120 & \shortstack{POPE: 3,000/3,000\\TVQA: 2,000} & 600 & spatial: 1,000 & attribute router; PEC: 3,000;\newline CHAIR: 500 \\
\addlinespace[3pt]
\multicolumn{6}{@{}l}{\textbf{Q1-only scale replications}} \\
LLaVA-NeXT-34B & 120 & --- & --- & --- & --- \\
Qwen2.5-VL-32B & 120 & --- & --- & --- & --- \\
\bottomrule
\end{tabular}
\end{table}

\subsection{Splits, interventions, and inference}
\label{app:experimental_protocol}
EAB uses the same 120 examples for every checkpoint: 48 examples fit the
binary tuned lens and determine a single checkpoint-level EAB, which is then
frozen. The remaining 72 queries are completely disjoint and are used only to
validate post-boundary persistence and estimate the reported pre-/post-EAB
effect ratios. The primary layer intervention deletes the visual residual
contribution and rescales every edit to equal RMS across depth, separating
layer position from perturbation magnitude. No evaluation query is used to
select or revise the boundary.

For Q2, the primary counterfactual mean-ablates visual-token
representations while holding the prompt and claim fixed. All claims or pairs associated with one
image remain in the same split and in the same bootstrap or permutation
cluster. PEC is evaluated at the calibration-selected operating point, and
PEC presence comparisons use the cached image-cluster McNemar audit. Thus,
no evaluation image contributes to threshold selection, and no test-instance
label is available to the router. Question category is taken from dataset
metadata when available and otherwise assigned by deterministic rules over
question form that are fixed before scoring.
Unless otherwise noted, image preprocessing, prompt construction,
tokenization, and answer extraction follow the released checkpoint defaults
and established protocols used in prior multimodal hallucination evaluations
\citep{li2023pope,wang2023amber,guan2024hallusion}; deviations specific to our
interventions are described in this section.

\subsection{Reproducibility checklist}
\label{app:reproducibility_checklist}
Table~\ref{tab:reproducibility} connects the frozen choices used by the main
claims to the records used for verification. The checklist is
organized to distinguish environment control, experimental design, and
artifact integrity.

\begin{table}[H]
\centering
\scriptsize
\setlength{\tabcolsep}{3.2pt}
\renewcommand{\arraystretch}{1.16}
\caption{\textbf{Reproducibility checklist.} Frozen policies and the records
used to verify the reported experiments and artifacts, supporting
reconstruction of the reported evaluation pipeline.}
\label{tab:reproducibility}
\begin{tabularx}{\textwidth}{@{}>{\raggedright\arraybackslash}p{0.15\textwidth}>{\raggedright\arraybackslash}p{0.31\textwidth}>{\raggedright\arraybackslash}X@{}}
\toprule
Item & Frozen value / policy & Reproduction record \\
\midrule
\multicolumn{3}{@{}l}{\textbf{Environment and determinism}} \\
\textbf{Hardware} & \(2\times\) NVIDIA A100-SXM4-80GB; CUDA driver 550.54.14 & one checkpoint per GPU; no model sharding for reported base runs \\
\textbf{Software} & Python; PyTorch 2.8.0, Transformers 4.51.0, NumPy 1.26.4, SciPy 1.17.1 & versioned dependency and environment records \\
\textbf{Randomness} & analysis seed 0; bootstrap seeds 0/1/2 where repeated & deterministic sample designs are cached \\
\addlinespace[3pt]
\multicolumn{3}{@{}l}{\textbf{Experimental design and inference}} \\
\textbf{EAB split} & 120 examples/checkpoint: 48 boundary calibration, 72 completely disjoint evaluation; one checkpoint-level EAB is frozen & cached split assignments and boundary-analysis records \\
\textbf{Counterfactual} & visual-token mean ablation for the reported Q2 comparisons & fixed prompt and claim \\
\textbf{Question category} & dataset metadata; deterministic question-form rules otherwise & mapping frozen before scoring; no reference answer or correctness input \\
\textbf{Uncertainty} & 2,000 paired image-cluster bootstrap resamples & resample images with all associated queries/pairs \\
\textbf{Hypothesis tests} & 10,000 cluster sign flips with \((b+1)/(B+1)\) correction; cached McNemar audit for PEC presence & test type specified in Appendix~\ref{app:significance_audit} \\
\textbf{Attribute router} & five-fold image-disjoint cross-validation & cached fold assignments and selection records; labels are never test-time inputs \\
\addlinespace[3pt]
\multicolumn{3}{@{}l}{\textbf{Artifact regeneration and integrity}} \\
\textbf{Summary refresh} & cached result files only & deterministic aggregation of cached outputs \\
\textbf{Paper artifacts} & vector PDF/SVG plus 400-dpi PNG & deterministic figure generation and manuscript compilation \\
\textbf{Integrity} & SHA-256 over result summaries and EAB caches & versioned checksum manifest \\
\textbf{Regression suite} & cache loaders, statistics, geometry, EAB, and figures & automated validation tests \\
\textbf{Baseline scope} & matched-next-token audit only for ICD/DeCo & no claim of reproducing full decoding stacks \\
\bottomrule
\end{tabularx}
\end{table}

\section{Checkpoint-Level Evidence for EAB and ACP}
\label{app:q1_additional}

This appendix supplies the checkpoint-level evidence behind
Section~\ref{sec:eab}. Section~\ref{app:eab_summary} expands the aggregate
EAB result in Figure~\ref{fig:eab_discovery};
Section~\ref{app:scale_comparisons} separates absolute- and relative-depth
changes across checkpoint sizes; Section~\ref{app:acp_boundary} tests the
stronger interpretation of ACP as an irreversibility boundary; and
Section~\ref{app:spatial_leverage} distinguishes spatial localization from
causal leverage.

\subsection{Per-checkpoint EAB summary}
\label{app:eab_summary}
Table~\ref{tab:eab_checkpoint_summary} reports the checkpoint-level boundaries
frozen from calibration queries and the completely held-out effect ratios
behind Figure~\ref{fig:eab_discovery}.
The absolute boundary varies across architectures, while every checkpoint
shows a large equal-RMS leverage drop on fully held-out evaluation queries.
This table is the numerical basis for
the cross-family range reported in Section~\ref{sec:eab}.

\begin{table}[H]
\centering
\footnotesize
\setlength{\tabcolsep}{3.5pt}
\renewcommand{\arraystretch}{1.12}
\caption{\textbf{The operator-specific EAB pattern replicates across all six checkpoints.} One
checkpoint-level boundary is frozen from calibration queries before evaluation;
relative EAB is that boundary divided by language-model depth. The resolution
column summarizes the cached boundary audit, whereas the pre/post ratios and
confidence intervals use the completely disjoint evaluation queries.}
\label{tab:eab_checkpoint_summary}
\begin{tabularx}{\textwidth}{@{}>{\raggedright\arraybackslash}p{0.21\textwidth}*{5}{>{\centering\arraybackslash}p{0.105\textwidth}}>{\centering\arraybackslash}X@{}}
\toprule
Checkpoint & \shortstack{LM\\depth} & \shortstack{Frozen\\EAB} & \shortstack{Relative\\EAB} & \shortstack{Boundary\\resolution} & \multicolumn{2}{c}{Equal-RMS pre/post effect} \\
\cmidrule(l){6-7}
& & & & & Ratio & 95\% CI \\
\midrule
LLaVA-1.5-7B & 32 & 12 & \(.375\) & 99.2\% & \(13.88\times\) & \([11.52, 16.49]\) \\
LLaVA-1.5-13B & 40 & 13 & \(.325\) & 99.2\% & \(13.78\times\) & \([10.98, 17.97]\) \\
LLaVA-NeXT-34B & 60 & 33 & \(.550\) & 100\% & \(26.10\times\) & \([22.24, 30.74]\) \\
\addlinespace[2pt]
Qwen2.5-VL-7B & 28 & 16 & \(.571\) & 100\% & \(17.54\times\) & \([15.22, 20.42]\) \\
Qwen2.5-VL-32B & 64 & 40 & \(.625\) & 97.5\% & \(18.24\times\) & \([14.89, 22.59]\) \\
\addlinespace[2pt]
InternVL3.5-8B & 36 & 18 & \(.500\) & 100\% & \(24.95\times\) & \([22.02, 28.50]\) \\
\bottomrule
\end{tabularx}
\end{table}

\subsection{Checkpoint-scale comparisons}
\label{app:scale_comparisons}
Table~\ref{tab:scale_shift} separates absolute layer shifts from relative-depth shifts.
The cached paired scale audit gives a \(+22.64\)-layer shift for
Qwen2.5-VL 7B\(\rightarrow\)32B, while its relative EAB shift is unresolved.
These audit estimates are distinct from subtracting the frozen
checkpoint-level indices in Table~\ref{tab:eab_checkpoint_summary}.
The LLaVA-13B\(\rightarrow\)NeXT-34B audit shows a later EAB in relative
depth. This comparison also changes model generation and language backbone,
so it combines scale and architecture differences.


\begin{table}[H]
\centering
\scriptsize
\setlength{\tabcolsep}{1.5pt}
\renewcommand{\arraystretch}{1.18}
\caption{\textbf{Checkpoint-scale audit.} (Shifts are larger minus smaller;
bracketed values are 95\% image-cluster confidence intervals.)}
\label{tab:scale_shift}
\begin{tabularx}{0.99\textwidth}{@{}>{\raggedright\arraybackslash}p{0.27\textwidth}*{2}{>{\centering\arraybackslash}p{0.18\textwidth}}>{\raggedright\arraybackslash}X@{}}
\toprule
\shortstack[l]{Comparison\\(depths; resolved)} & \shortstack{\(\Delta\)EAB\\(layers)} & \shortstack{\(\Delta\) relative\\EAB} & Interpretation \\
\midrule
\shortstack[l]{LLaVA-1.5-13B\\\(\rightarrow\) LLaVA-NeXT-34B\\40\(\rightarrow\)60 layers;\\119/120 resolved} & \shortstack{\(+17.72\)\\\([16.25,19.12]\)} & \shortstack{\(+.166\)\\\([.131,.199]\)} & later EAB in relative depth \\
\addlinespace[3pt]
\shortstack[l]{Qwen2.5-VL-7B\\\(\rightarrow\) Qwen2.5-VL-32B\\28\(\rightarrow\)64 layers;\\117/120 resolved} & \shortstack{\(+22.64\)\\\([21.67,23.66]\)} & \shortstack{\(+.019\)\\\([-.007,.045]\)} & relative EAB shift unresolved \\
\bottomrule
\end{tabularx}
\end{table}

\subsection{ACP as an answer-readout coordinate}
\label{app:acp_boundary}
Raw and held-out tuned lenses agree that candidate readouts become more
reliable with depth. Query-level interventions can still flip outputs after
the readout has stabilized. In the controlled
object/attribute/relation audit of 96 triplets per model, the tested windowed
edit corrects two LLaVA-7B baseline errors and zero errors for LLaVA-13B,
Qwen2.5-VL-7B, and InternVL\@. The sparse corrections under this controlled
audit limit claims about a general correction window. ACP records readout
stabilization; intervention tests separately establish causal reach.

\subsection{Spatial localization and intervention leverage}
\label{app:spatial_leverage}
\begin{figure}[H]
\centering
\includegraphics[width=\linewidth]{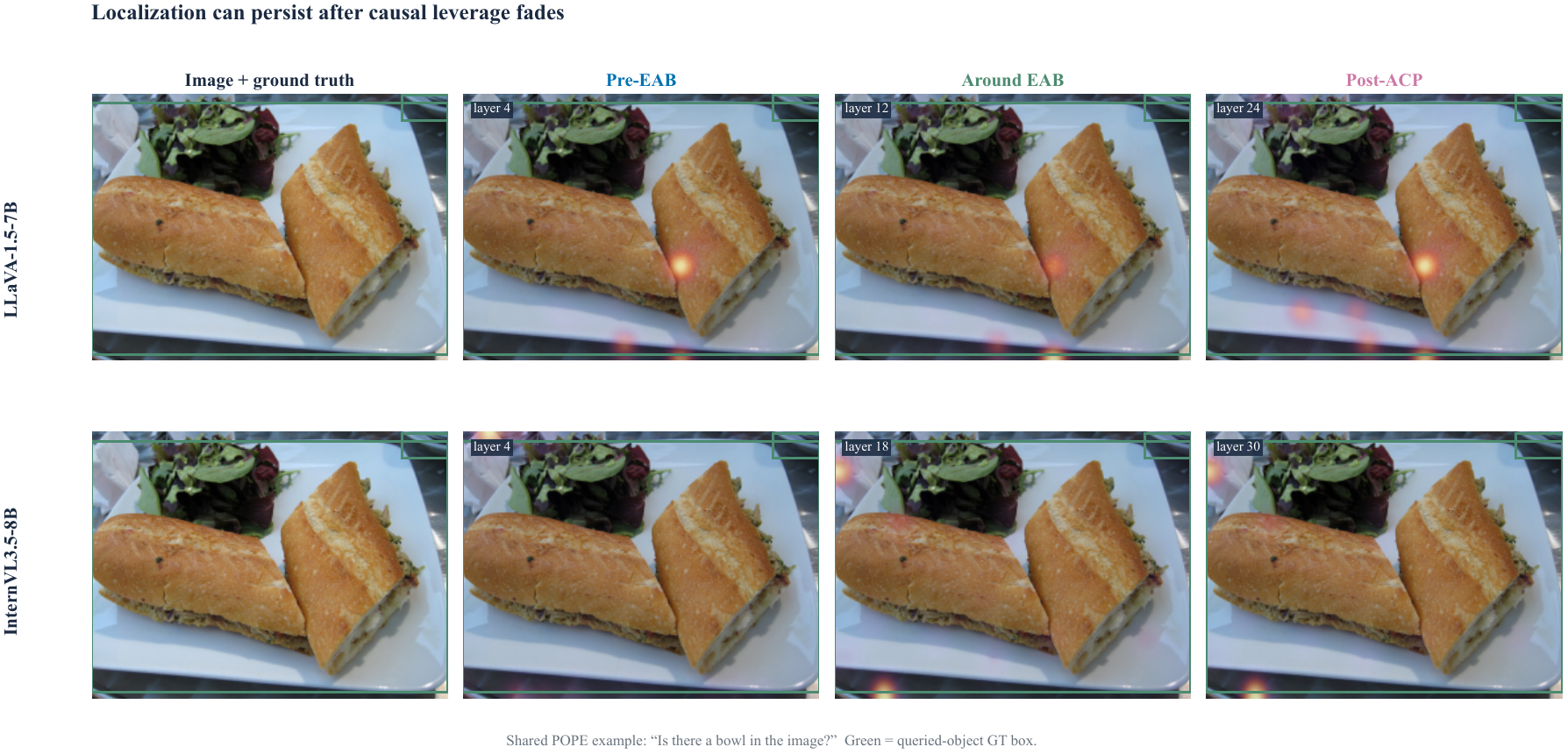}
\caption{\textbf{Spatial localization can persist after direct visual-residual leverage fades.}
Head-averaged last-query attention on one shared example; green marks the queried-object region.}
\label{fig:attention_depth}
\end{figure}

The model-level intervention effect and a token-attention map answer different questions.
Figure~\ref{fig:attention_depth} shows one shared POPE example before EAB, near EAB, and after ACP: layers \(4/12/24\) for LLaVA-1.5-7B and \(4/18/30\) for InternVL3.5-8B.
Both models still attend near the queried bowl at late depth even after the aggregate equal-RMS intervention effect has declined.
The attention map shows where the model attends, while the intervention-effect
measurement in Section~\ref{sec:eab} tests leverage. This single case
illustrates the distinction under these conditions without estimating its prevalence.

\section{EAB Controls and Boundary Interpretation}
\label{app:boundary_robustness}

This appendix explains what the equal-RMS control and frozen-boundary
evaluation establish. The numerical evidence is in
Table~\ref{tab:eab_checkpoint_summary}; the interpretation depends on both
the intervention operator and the boundary rule and does not imply a
universal architectural phase transition.

\subsection{Matched perturbation strength and operator scope}
\label{app:operator_robustness}
Equal-RMS edits at every layer control for variation in perturbation norm.
Under this control, the primary direct visual-residual edit still has a
large held-out pre-/post-EAB effect ratio in every checkpoint. Matching
norm does not equate different edit directions, targeted representations,
or downstream propagation paths. EAB therefore characterizes the response
to the tested operator.

The manuscript does not report a quantitative cross-operator boundary
agreement statistic. In particular, the primary Q2 mean-ablation results
do not by themselves establish that mean, zero, shuffle, and residual
interventions share an EAB. The six-checkpoint replication establishes
the depth pattern for the specified Q1 edit, with separate calibration for
each checkpoint.

\subsection{Boundary location and effect-size uncertainty}
\label{app:boundary_sensitivity}
The primary rule requires a three-layer local mean and a remaining-suffix 75th
percentile below \(10\%\) of the checkpoint-level calibration trajectory peak.
The selected boundary is then frozen and applied unchanged to the completely
disjoint evaluation queries.
The confidence intervals in Table~\ref{tab:eab_checkpoint_summary}
resample held-out images while retaining that boundary by construction. They quantify
uncertainty in the effect contrast, not in the calibration-selected layer.
Likewise, the resolution column records whether the cached boundary audit
returns an index; it does not measure the precision of that index.

The reported results do not quantify how the boundary moves under
\(5\%\), \(15\%\), or \(20\%\) cutoffs, alternative suffix statistics,
or persistence lengths. The primary rule gives a reproducible division
of the effect trajectory, and the independent evaluation shows that its
early and late regions differ substantially. A smooth decay can also
satisfy this rule. Distinguishing a change point from smooth decay would
require an explicit model comparison; no such comparison is reported here.

\section{Question-Category Diagnostics for Q2}
\label{app:q2_additional}

This appendix supports the interpretation in Section~\ref{sec:geometry} by
testing two simpler alternatives: that presence errors can be identified from
spatial attention alone, or that prompt form induces a checkpoint-invariant
correction. The attention audit evaluates localization; the prompt audit
motivates re-probing after neutralization, while keeping their evidential
roles explicitly separate in practice.

\subsection{Aggregate spatial-attention diagnostic}
\label{app:attention_diagnostic}
\begin{figure}[H]
\centering
\includegraphics[width=\linewidth]{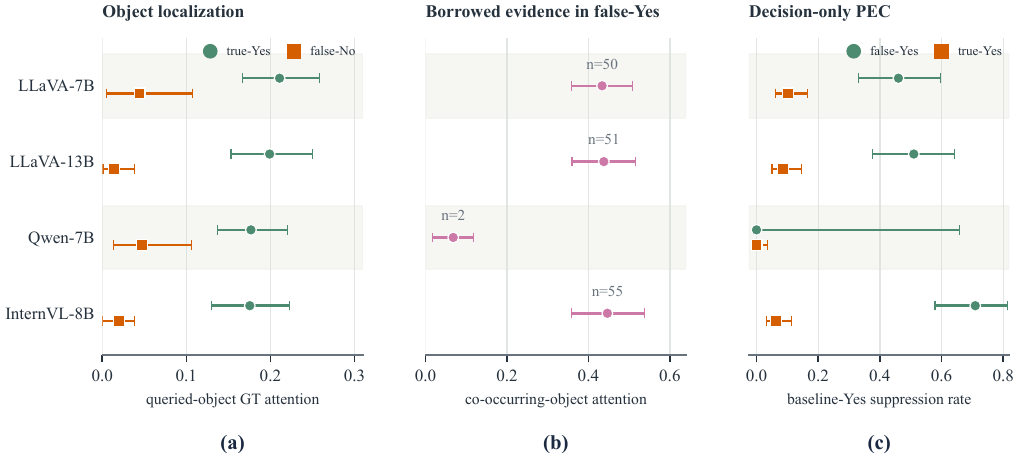}
\caption{\textbf{Aggregate spatial-attention audit across four base checkpoints.}
Localization does not by itself identify whether a score intervention will improve the answer.}
\label{fig:attention_summary}
\end{figure}

Figure~\ref{fig:attention_summary} uses a fixed 300-example subset per base checkpoint.
It compares queried-object attention for true-Yes decisions and false-No misses,
measures attention to co-occurring objects among false-Yes decisions, and reports
PEC suppression for false-Yes and true-Yes cases.
The comparison examines whether localization distinguishes correct and
incorrect decisions. The observed maps describe
where the model attends but yield no stable correction rule for downstream routing
across checkpoints and decision types in this audit; we therefore exclude them
from router inputs.

\subsection{Prompt presupposition is checkpoint dependent}
\label{app:presupposition_diagnostic}
\begin{figure}[H]
\centering
\includegraphics[width=\linewidth]{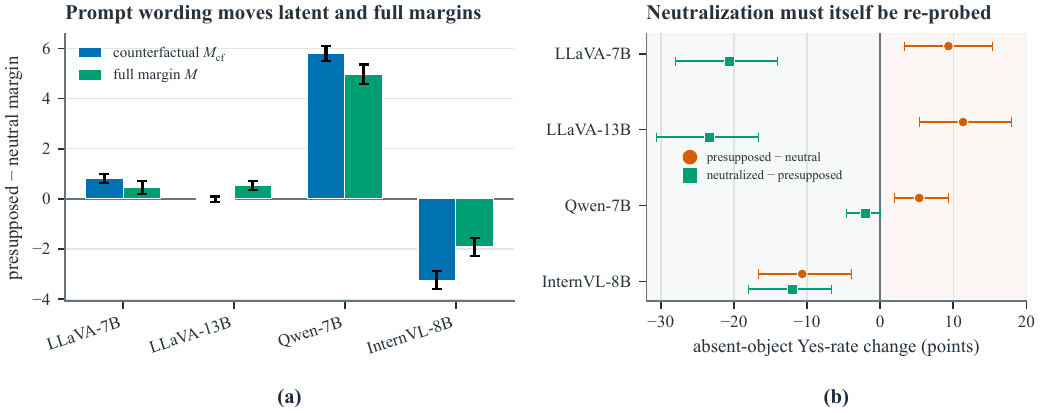}
\caption{\textbf{Prompt presupposition and neutralization are checkpoint-dependent operators.}}
\label{fig:presupposition_scope}
\end{figure}

Figure~\ref{fig:presupposition_scope} uses 150 absent-object examples from each model's frozen POPE subset.
Presupposition changes false-Yes rates by \(+5.3\) to \(+11.3\) points in the two LLaVA checkpoints and Qwen, but by \(-10.7\) points in InternVL; neutralization is likewise strong for some checkpoints and nearly null for Qwen.
The effect therefore depends on the checkpoint. We consequently
neutralize presupposed wording and then re-probe the claim set before
selecting a correction (Appendix~\ref{app:prompt_handling}).

\section{Router Calibration and Routing Protocol}
\label{app:verifier_details}

We describe how to select and evaluate the routing rules in
Section~\ref{sec:verifier} and Figure~\ref{fig:verifier_pipeline}. Calibration uses images separate from those
used for evaluation, with no image overlap throughout the routing protocol.
The available operators depend on the evaluated
model--task combinations in Table~\ref{tab:models_data}.

\subsection{Image-disjoint calibration}
\label{app:image_disjoint_calibration}
We divide images into five folds, keeping all claims from an image in
the same fold. In each rotation, we use four folds to calibrate the
attribute router and evaluate it on the remaining fold. We first estimate
ranking disagreement, counterfactual score spread, and sign consistency
on the calibration data. We then select the threshold and fix the rule
before evaluation. Each image receives one prediction from a rule
selected without that image. Answer labels define the wrong/helpful
alignment groups for aggregate analysis; the router receives no such
alignment labels at test time.

For PEC, we select the layer \(\ell^\star\) and coefficient \(\alpha\)
on calibration data. We keep both fixed during evaluation and transfer
to other tasks. The transfer results therefore measure how the original
configuration behaves on a new task without any task-specific retuning afterward.

\subsection{Question-category assignment}
\label{app:category_assignment}
We use the benchmark's category metadata when available. Otherwise, we
apply deterministic rules over question form. We fix the mapping before
model scoring and exclude reference answers, predictions, and correctness
from its inputs. The resulting category, independent of model output signals,
determines which calibrated operators the router can consider.

\subsection{Prompt handling}
\label{app:prompt_handling}
For presupposed presence questions, we first rewrite the question into a
neutral claim and then apply the usual presence probe. We require
calibration support for any subsequent score change.
Figure~\ref{fig:presupposition_scope} shows why this step is needed:
the effect of neutralization differs across checkpoints. Re-probing tests
the normalized claim for scoring before the router selects an action.

\subsection{Inference computation}
\label{app:inference_cost}
For a set of \(K\) claims, the basic subtraction score requires \(K\)
original-condition scores and \(K\) counterfactual-condition scores,
compared with \(K\) scores for the original model. Additional prompt
or stability probes can add evaluations. The routing decision and score
blend operate on the resulting scores. Preserving \(M\) after probing
preserves the prediction but does not recover computation already spent
on the counterfactual, even when the original score is preserved.

This count describes the score evaluations implied by the method, not
measured latency or FLOPs. Batching, prefix reuse, visual-token count,
and prompt re-probing affect the actual runtime. We do not report a
wall-clock comparison with the original model or external decoding methods.

\section{Extended Evaluation and Statistical Audits}
\label{app:extended_evaluation}

This appendix provides the numerical evidence behind
Section~\ref{sec:experiments}. It expands the routed attribute and targeted-
generation outcomes, reports the calibrated PEC operating range and transfer
failures, and gives the image-cluster-aware significance audit. The
final subsection states the limited reproduction scope of the secondary
baselines without extending beyond the reproduced configurations.

\subsection{Detailed attribute-routing outcomes}
\label{app:routing_outcomes}
\begin{figure}[H]
\centering
\includegraphics[width=\linewidth]{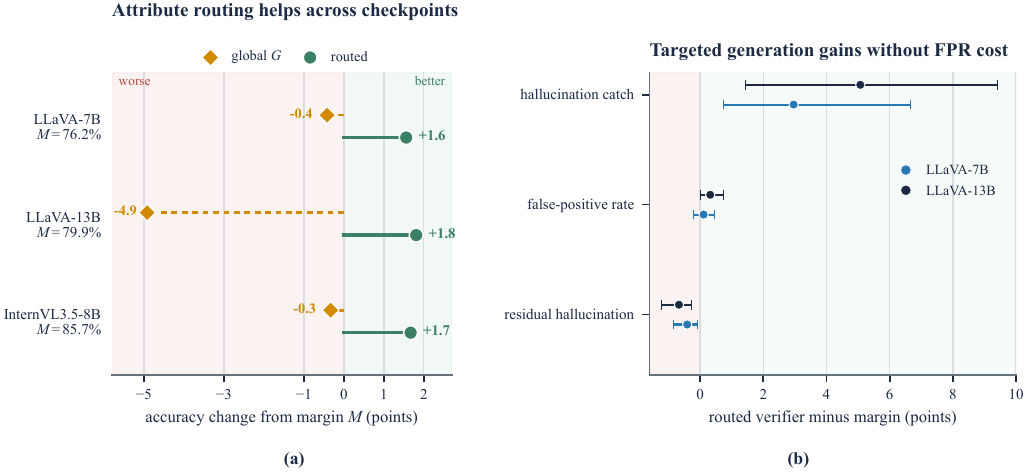}
\caption{\textbf{Detailed attribute-routing and targeted-generation results.}
(a) The routed score improves full-set held-out attribute accuracy while
global subtraction is neutral or harmful. (b) Point estimates show higher
hallucination catch, small false-positive changes, and lower residual
hallucination. Horizontal bars are 95\% image-cluster intervals; paired tests
are in Table~\ref{tab:paired_significance}.}
\label{fig:verifier_results_detail}
\end{figure}

Figure~\ref{fig:verifier_results_detail} expands the attribute and targeted-
generation panels summarized in the main-text Figure~\ref{fig:selective_routing_results}.
It shows the absolute held-out pair accuracies behind the reported routing
gains and separates hallucination catch, false-positive rate, and residual
hallucination for the two targeted-generation checkpoints, while preserving
the distinction between these downstream failure modes.

\subsection{PEC operating range and out-of-domain transfer}
\label{app:pec_transfer}
\begin{figure}[H]
\centering
\includegraphics[width=\linewidth]{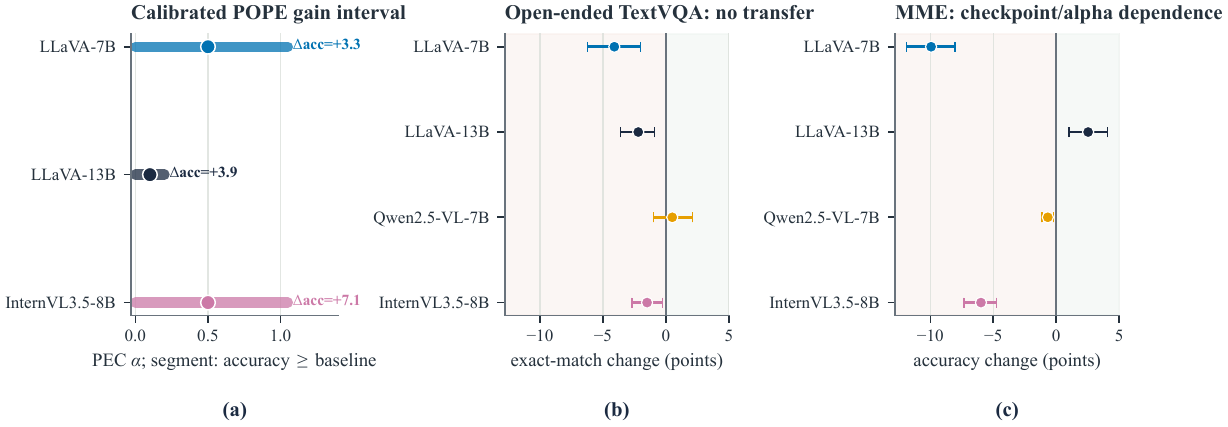}
\caption{\textbf{PEC has a calibrated presence operating range but does not transfer universally.}
Segments mark POPE gain regions; transfer whiskers are 95\% paired intervals.}
\label{fig:pec_scope}
\end{figure}

Figure~\ref{fig:pec_scope} shows the \(\alpha\) interval over which POPE accuracy is at least the baseline and applies each frozen configuration to TextVQA and MME\@.
Open-ended TextVQA loses accuracy for three of four checkpoints, and full MME changes range from \(-9.9\) to \(+2.5\) points.
Each transfer test uses the original frozen configuration without any task-specific retuning.
On the lower-headroom POPE-random split, the same correction changes accuracy by \(+0.8\), \(+0.8\), and \(-0.3\) points for LLaVA-7B, LLaVA-13B, and InternVL, further limiting the scope of the presence result.

\begin{figure}[H]
\centering
\includegraphics[width=\linewidth]{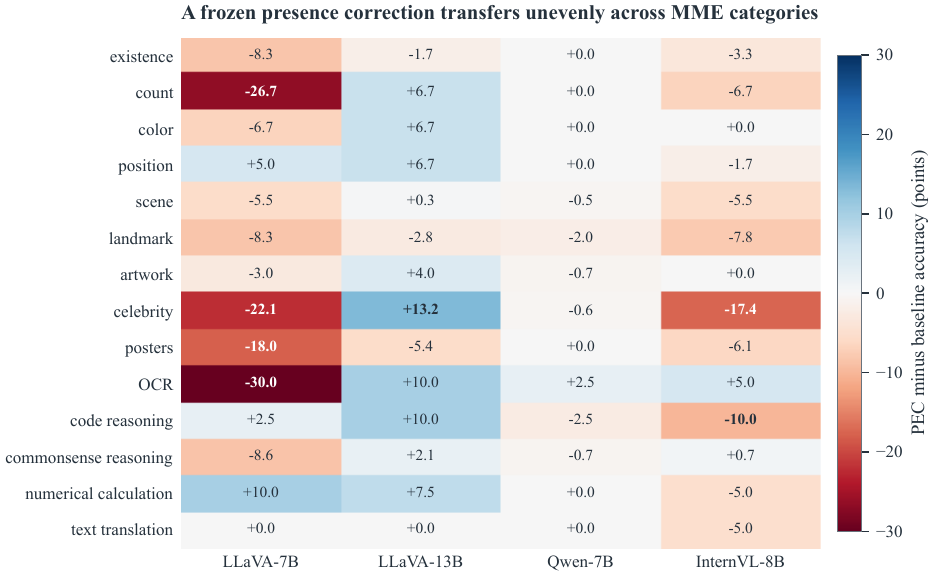}
\caption{\textbf{Frozen PEC transfer is heterogeneous across MME categories and checkpoints.}}
\label{fig:mme_categories}
\end{figure}

Figure~\ref{fig:mme_categories} decomposes the frozen MME scores.
LLaVA-7B loses most on OCR, counting, celebrity, and poster recognition while improving numerical calculation and position; LLaVA-13B improves several categories but loses on posters and landmarks.
Qwen changes little in aggregate under its mild configuration, whereas InternVL has substantial celebrity and code-reasoning losses.

\subsection{Paired significance audit}
\label{app:significance_audit}
For row-level accuracy contrasts, let \(d_i=A_i-B_i\in\{-1,0,1\}\).
The reported effect is \(100\bar d\) in percentage points.
A 95\% percentile interval uses 2,000 paired bootstrap resamples of image clusters.
For PEC presence comparisons, \(p\)-values use the cached McNemar audit
(Appendix~\ref{app:experimental_protocol}). For the other accuracy contrasts,
the two-sided permutation test flips all \(d_i\) from the same image together,
uses 10,000 Monte Carlo draws, and applies the finite-sample correction
\((b+1)/(B+1)\).
Targeted-generation residual hallucination is a ratio, so only its cluster-bootstrap interval is reported.
Table~\ref{tab:paired_significance} reports these tests for attribute routing,
presence and inverse-spatial subtraction, targeted generation, and
calibrated PEC presence correction.
Direct-contact confidence intervals are in Table~\ref{tab:category_results}.
The LLaVA-7B hallucination-catch increase has a positive bootstrap interval
but remains unresolved by the paired test (\(p=.1314\)); we report both
summaries without treating them as interchangeable.

\begingroup
\setlength{\tabcolsep}{3.2pt}
\renewcommand{\arraystretch}{1.12}

\begin{longtable}{@{}>{\raggedright\arraybackslash}p{0.21\linewidth}
>{\raggedright\arraybackslash}p{0.20\linewidth}
>{\centering\arraybackslash}p{0.07\linewidth}
>{\centering\arraybackslash}p{0.08\linewidth}
>{\centering\arraybackslash}p{0.21\linewidth}
>{\centering\arraybackslash}p{0.10\linewidth}@{}}

\caption{\textbf{Paired significance audit.} Effects are \(A-B\)
percentage points; inference is image-cluster aware. (A dash in the \(p\)
column indicates a ratio outcome for which only a cluster-bootstrap interval
is reported.)}
\label{tab:paired_significance}\\
\toprule
Outcome & Checkpoint & \(n\) & \shortstack{Effect\\(pp)} & 95\% CI & \(p\) \\
\midrule
\endfirsthead
\multicolumn{6}{c}{\tablename\ \ref{tab:paired_significance}\ (continued)}\\
\toprule
Outcome & Checkpoint & \(n\) & \shortstack{Effect\\(pp)} & 95\% CI & \(p\) \\
\midrule
\endhead
\midrule
\multicolumn{6}{r}{\textit{Continued on next page}} \\
\bottomrule
\endfoot
\bottomrule
\endlastfoot
\multicolumn{6}{@{}l}{\textbf{Attribute subtraction (\(G-M\))}} \\
All attributes & LLaVA-1.5-7B & 2,378 & \(-0.42\) & \([-1.64, +0.84]\) & \(.5491\) \\
Hard attributes & LLaVA-1.5-7B & 521 & \(+13.44\) & \([+10.17, +16.89]\) & \(<10^{-4}\) \\
All attributes & LLaVA-1.5-13B & 2,378 & \(-4.92\) & \([-6.39, -3.36]\) & \(<10^{-4}\) \\
Hard attributes & LLaVA-1.5-13B & 508 & \(+7.68\) & \([+3.15, +12.01]\) & \(.0010\) \\
All attributes & InternVL3.5-8B & 600 & \(-0.33\) & \([-2.83, +2.17]\) & \(.8946\) \\
Hard attributes & InternVL3.5-8B & 73 & \(+17.81\) & \([+6.85, +28.77]\) & \(.0044\) \\
\addlinespace[2pt]
\multicolumn{6}{@{}l}{\textbf{Selective attribute routing (router-\(M\))}} \\
Full attributes & LLaVA-1.5-7B & 2,378 & \(+1.56\) & \([+0.97, +2.19]\) & \(<10^{-4}\) \\
Full attributes & LLaVA-1.5-13B & 2,378 & \(+1.81\) & \([+1.18, +2.44]\) & \(<10^{-4}\) \\
Full attributes & InternVL3.5-8B & 600 & \(+1.67\) & \([+0.67, +2.83]\) & \(.0058\) \\
\addlinespace[2pt]
\multicolumn{6}{@{}l}{\textbf{Presence and relation subtraction (\(G-M\))}} \\
TVQA presence & LLaVA-1.5-7B & 1,500 & \(-0.80\) & \([-1.60, 0.00]\) & \(.0656\) \\
TVQA presence & LLaVA-1.5-13B & 1,500 & \(-0.67\) & \([-1.27, -0.07]\) & \(.0508\) \\
TVQA presence & Qwen2.5-VL-7B & 1,500 & \(-1.93\) & \([-2.87, -1.07]\) & \(.0002\) \\
TVQA presence & InternVL3.5-8B & 1,500 & \(+0.07\) & \([-0.53, +0.73]\) & \(1.0000\) \\
Inverse spatial & LLaVA-1.5-7B & 1,000 & \(-2.30\) & \([-3.90, -0.70]\) & \(.0070\) \\
Inverse spatial & LLaVA-1.5-13B & 1,000 & \(-0.70\) & \([-2.30, +0.90]\) & \(.4749\) \\
Inverse spatial & Qwen2.5-VL-7B & 1,000 & \(-0.30\) & \([-1.20, +0.60]\) & \(.6434\) \\
Inverse spatial & InternVL3.5-8B & 1,000 & \(+0.40\) & \([-0.40, +1.20]\) & \(.4507\) \\
\addlinespace[2pt]
\multicolumn{6}{@{}l}{\textbf{Targeted generation (router-\(M\))}} \\
Hallucination catch & LLaVA-1.5-7B & 135 & \(+2.96\) & \([+0.74, +6.67]\) & \(.1314\) \\
False-positive rate & LLaVA-1.5-7B & 905 & \(+0.11\) & \([-0.22, +0.44]\) & \(1.0000\) \\
Hallucination catch & LLaVA-1.5-13B & 138 & \(+5.07\) & \([+1.45, +9.42]\) & \(.0168\) \\
False-positive rate & LLaVA-1.5-13B & 939 & \(+0.32\) & \([0.00, +0.75]\) & \(.2449\) \\
Residual hallucination & LLaVA-1.5-7B & 1,040 & \(-0.41\) & \([-0.83, -0.09]\) & --- \\
Residual hallucination & LLaVA-1.5-13B & 1,077 & \(-0.67\) & \([-1.22, -0.26]\) & --- \\
\addlinespace[2pt]
\multicolumn{6}{@{}l}{\textbf{Presence correction (PEC-baseline)}} \\
POPE accuracy & LLaVA-1.5-7B & 3,000 & \(+3.30\) & \([+2.23, +4.40]\) & \(8.47\!\times\!10^{-9}\) \\
POPE accuracy & LLaVA-1.5-13B & 3,000 & \(+3.90\) & \([+2.67, +5.17]\) & \(1.26\!\times\!10^{-9}\) \\
POPE accuracy & InternVL3.5-8B & 3,000 & \(+7.07\) & \([+5.63, +8.50]\) & \(4.84\!\times\!10^{-23}\) \\
\end{longtable}
\endgroup

\subsection{Secondary baseline scope}
\label{app:secondary_baselines}
The ICD and DeCo audits evaluate their released scoring formulas through a
matched one-token interface. Full decoding additionally involves
method-specific control flow and stopping rules. These audits check score
behavior under shared conditions; the main comparison evaluates \(M\),
universal \(G\), and routing. End-to-end competitiveness with the original
decoding methods remains untested.

\section{Qualitative Examples of Intervention Outcomes}
\label{app:qualitative}

This appendix illustrates the mechanisms summarized in
Sections~\ref{sec:geometry} and~\ref{sec:experiments}. The selected examples
explain decision changes and localization errors; they are not used to
estimate prevalence or calibrate the router, while keeping their role
illustrative rather than evaluative.

\subsection{Attribute rescue, damage, and preservation}
\label{app:qualitative_attributes}
\begin{figure}[H]
\centering
\includegraphics[width=\linewidth]{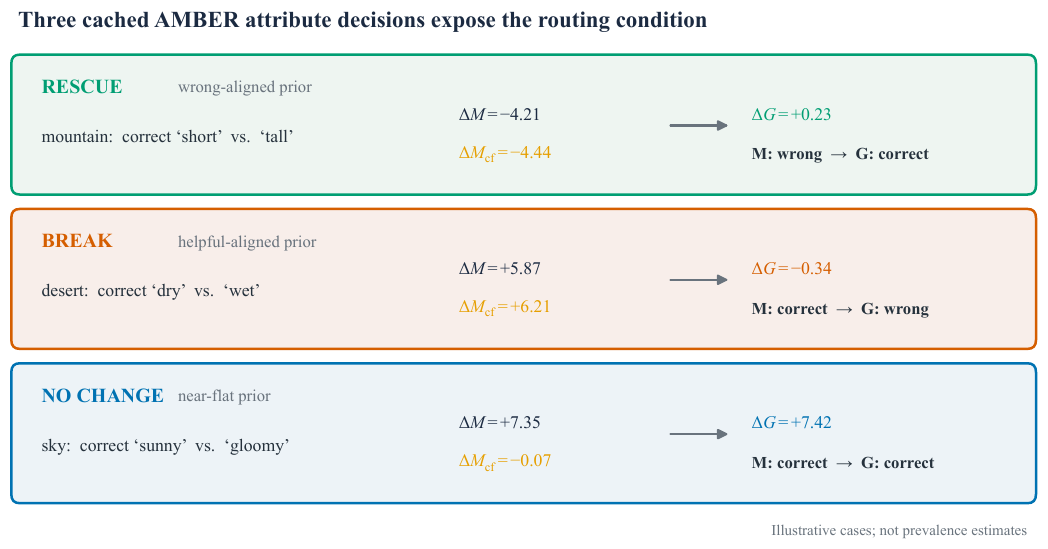}
\caption{\textbf{Concrete attribute decisions expose the routing condition.}
Counterfactual subtraction rescues, damages, or preserves a ranking according
to the relative preference it removes. These examples illustrate the
mechanism and are not prevalence estimates.}
\label{fig:attribute_cases_detail}
\end{figure}

Figure~\ref{fig:attribute_cases_detail} instantiates
Eq.~\ref{eq:shear_identity} on three cached LLaVA-7B AMBER pairs,
complementing the schematic cases in Figure~\ref{fig:claim_geometry}b. The mountain
pair is rescued by subtraction, the desert pair loses a correct decision,
and the sky pair preserves its ranking. The removed preference explains
each outcome.

\Needspace{8\baselineskip}
\subsection{Presence localization cases}
\label{app:qualitative_presence}
\begin{figure}[H]
\centering
\includegraphics[width=\linewidth]{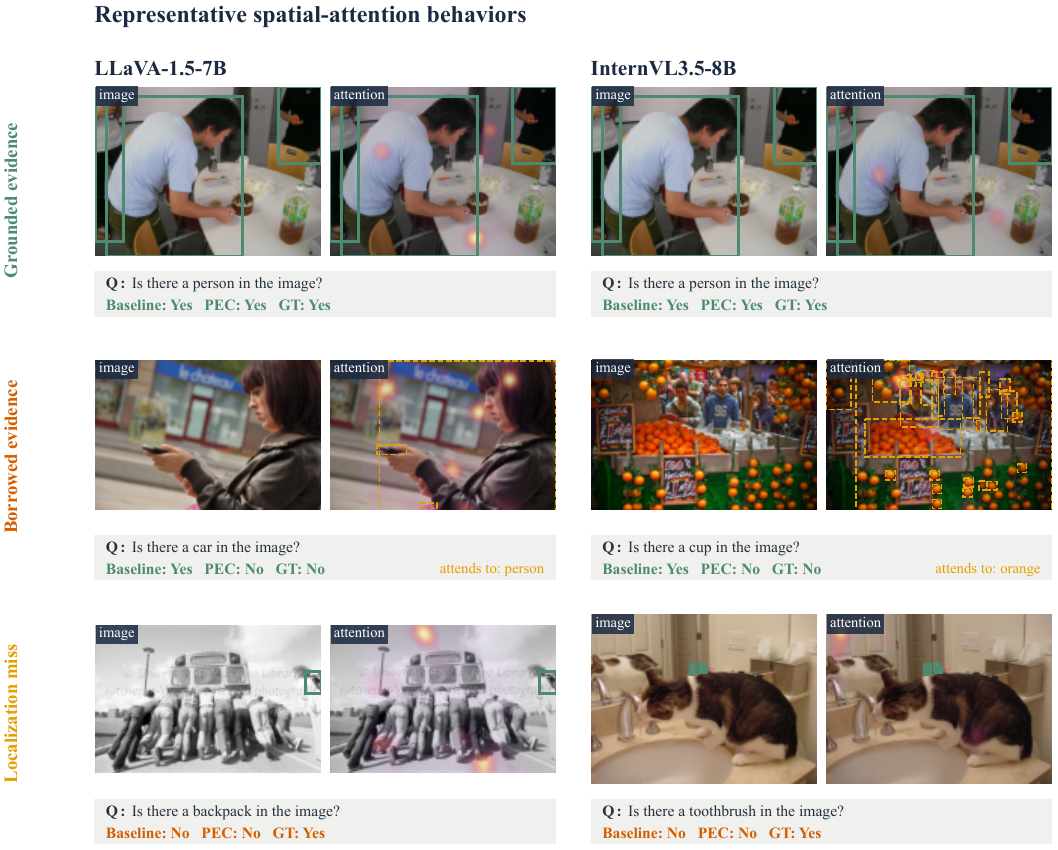}
\caption{\textbf{Representative grounded, borrowed-evidence, and localization-miss cases.}}
\label{fig:attention_cases}
\end{figure}

Figure~\ref{fig:attention_cases} shows grounded, borrowed-evidence, and localization-miss cases.
Solid green boxes mark the queried object; dashed amber boxes mark present co-occurring objects.
The cases clarify why localization is not itself an intervention rule:
attention may align with the queried object, borrow support from a co-occurring
object, or miss the relevant region. PEC changes the logit-space decision, not
the displayed attention map, and these examples are not used as PEC features.

\end{document}